\documentclass[letterpaper,10pt,conference]{ieeeconf}  

\IEEEoverridecommandlockouts                              

\usepackage{mymacros}
\usepackage{times} 
\usepackage{graphicx}

\usepackage[T1]{fontenc}
\usepackage{color}
\usepackage{hyperref}
\usepackage{pifont}

\let\labelindent\relax
\usepackage{enumitem}

\title{\LARGE \bf
Target-dependent UNITER: \\
A Transformer-Based Multimodal Language Comprehension Model \\
for Domestic Service Robots
}

\author{Shintaro Ishikawa$^{1}$ and Komei Sugiura$^{1}$
\thanks{$^{1}$Authors are with Keio University, 3-14-1 Hiyoshi, Kohoku, Yokohama, Kanagawa 223-8522, Japan.
        {\tt\small shin.0116@keio.jp, komei.sugiura@keio.jp}}
}

\begin{document}

\maketitle
\thispagestyle{empty}
\pagestyle{empty}

\begin{abstract}
Currently, domestic service robots have an insufficient ability to interact naturally through language.
This is because understanding human instructions is complicated by various ambiguities and missing information.
In existing methods, the referring expressions that specify the relationships between objects are insufficiently modeled.
In this paper, we propose Target-dependent UNITER, which learns the relationship between the target object and other objects directly by focusing on the relevant regions within an image, rather than the whole image.
Our method is an extension of the UNITER\cite{chen2020uniter}-based Transformer that can be pretrained on general-purpose datasets.
We extend the UNITER approach by introducing a new architecture for handling the target candidates.
Our model is validated on two standard datasets, and the results show that Target-dependent UNITER outperforms the baseline method in terms of classification accuracy. 
\end{abstract}
\section{Introduction
\label{sec:intro}
}

In our aging society, there is an increasing need for daily care and support. As a result, the shortage of home care workers has become a social problem, and domestic service robots (DSRs) that can physically assist people with disabilities are receiving increased attention. However, such robots currently have an insufficient ability to interact naturally through language.

In this study, we aim to develop a multimodal language understanding method that allows the DSRs to comprehend object-fetching instructions properly.
Specifically, given an instruction such as ``Grab the plastic bottle with red stripes and put it in the upper left box,'' it is desirable for the DSRs to recognize the target bottle among various others bottles and objects.

However, such human instructions are often ambiguous and do not always contain sufficient information. In the above example, if there are multiple ``bottles with red stripes'' in the scene, it is difficult to identify the correct target object from the given sentence alone.
Existing methods attempt to combine linguistic knowledge with visual knowledge by inputting a complete image containing the target object in addition to the verbal instruction\cite{magassouba2019understanding}. However, in such methods, the referring expressions that specify the relationships between the objects are insufficiently modeled. In addition, previous methods do not allow transfer knowledge from other tasks.

\begin{figure}[t]
    \centering
    \includegraphics[width=\linewidth]{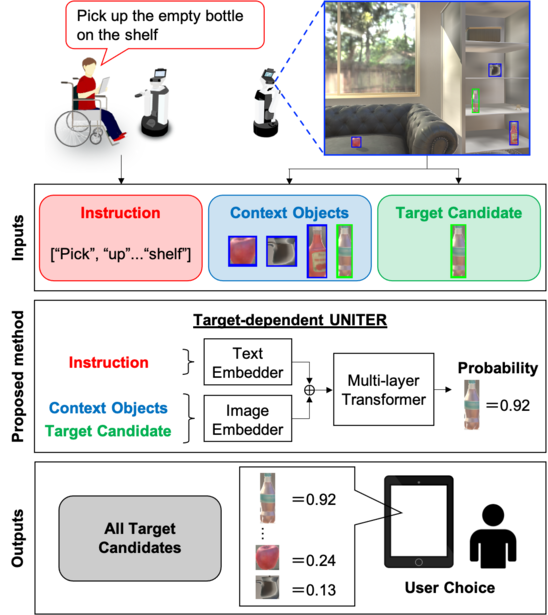}
    \caption{\small Target-dependent UNITER overview: Transformer is used to learn the relationship between instruction and objects.}
    \label{fig:overview}
\end{figure}

In this paper, we propose a Transformer\cite{vaswani2017attention}-based method that learns the relationship between the target object and other objects directly. Our method is based on UNITER\cite{chen2020uniter}. Our method, Target-dependent UNITER, focuses on their regions within an image, rather than the whole image.
Fig.~\ref{fig:overview} shows an overview of our method,
which consists of Image Embedder, Text Embedder, and Multi-layer Transformer modules.

The difference from existing methods is that our method is a UNITER-based Transformer model that can be pretrained on general-purpose datasets. Moreover, we extend the UNITER framework by introducing a new architecture for handling the visual features and location of the target candidate.
Our model can learn the relationships between all detected objects in the image and the text based on the attention mechanism in the Transformer, thus allowing the referring expressions regarding their spatial relationships to be fully comprehended. In addition, our method considers information about the target candidate to judge whether it is correct or not, something which UNITER is not capable of doing. A demonstration video is available at this URL\footnote{https://youtu.be/OOsPhoWFL4A}.

The main contributions of this paper are summarized as follows:
\begin{itemize}
 \item We combine the attention mechanism used in the UNITER with a general pretrained model for learning the relationship between an image and text in the field of object-manipulation instruction comprehension.
 \item We introduce a new structure for handling the target candidate in the UNITER framework.
\end{itemize}

\section{Related Work
\label{sec:related}
}

There have been many studies in the field of multimodal language processing. For example, \cite{mogadala2020trends} is a survey paper that explains ten representative tasks integrating vision and language, and discusses the associated problem formulations, methods, existing datasets, and evaluation metrics. It also compares the results obtained with the corresponding state-of-the-art methods.

This field can be divided into subfields depending on the combination of modalities. The main subfields are visual question answering (VQA)\cite{antol2015vqa}\cite{harabagiu2000experiments}, visual referring expression (VRE)\cite{krahmer2012computational}\cite{thomas2014meaning}, and visual entailment (VE)\cite{condoravdi2003entailment}\cite{bowman2015large}.

VQA aims to predict the correct answer given an image and a question. One pioneering study in this field is \cite{malinowski2014multi}, which views VQA as a visual Turing test and focuses on the implementation of the ability to semantically access visual information as humans do. In a recent study, \cite{agrawal2018don} stated that existing VQA models were heavily dependent on superficial correlations in the training data and were not accompanied by a visual basis for answering questions.

VRE aims to predict the target object given an image and a sentence containing a referring expression about that image. This field can be divided into two strands: comprehension and generation. A representative example of the former is \cite{yang2019cross}, which proposes two networks. The first is Cross-Modal Relationship Extractor, which uses a cross-modal attention mechanism to focus on the objects and relationships that are relevant to a given representation, and represents the extracted information as a visual relationship graph. The second is Gated Graph Convolutional Network, which fuses information from different modalities and computes the semantic context by propagating multimodal information in the structured relational graph. A representative work on the latter strand is \cite{luo2017comprehension}, which describes a comprehension module trained by human-generated representations for use as a ``critic'' of the referring expression generation, and the re-evaluation of the generated representations.
ABEN\cite{ogura2020alleviating} is an encoder--decoder model that generates instructions about objects in an image. This model performs the processing in the opposite direction to our method.

VE aims to determine whether the contents of an image and a sentence about the image are logically consistent or not. This field is relatively young and was first defined in \cite{xie2019visual}. This study created the SNLI-VE dataset for evaluating VE alongside various existing VQA models.

MTCM\cite{magassouba2019understanding} is the baseline method in this study. It differs from our method in that it uses LSTM\cite{hochreiter1997long} for language processing and VGG16\cite{simonyan2014very} for image processing.
MTCM-AB\cite{magassouba2020multimodal_01} is an extension of MTCM with ABN\cite{fukui2019attention}. This model develops a multimodal attention mechanism with attention branches and generates an attention map of objects in the image. MTCM-AB outperforms MTCM and \cite{hatori2018interactively} in terms of classification accuracy, again using LSTM for language processing and VGG16 for image processing.

Multimodal language understanding for fetching instructions (MLU-FI) is similar to VRE in that the goal is to ground objects in the image with respect to a referring expression. In the field of VRE, datasets using real images include RefCLEF\cite{kazemzadeh2014referitgame}, RefCOCO\cite{kazemzadeh2014referitgame}, and GuessWhat\cite{de2017guesswhat}, while those using synthetic images include CLEVR-Ref+\cite{liu2019clevr}.
Public datasets for MLU-FI include PFN-PIC\cite{hatori2018interactively} and WRS-PV\cite{ogura2020alleviating}, both of which consist of images and fetching instructions for objects in the images. The PFN-PIC dataset contains real images with objects scattered in four boxes, whereas the WRS-PV dataset contains images collected in the standard simulator of World Robot Summit Partner Robot Challenge\cite{okada2019competitions} by DSRs moving around the room.

Our method is based on the UNITER\cite{chen2020uniter} framework, but differs from UNITER in that it introduces a new structure for handling target candidates.
Similar to UNITER, ViLBERT\cite{lu2019vilbert} is a model for learning image--text embeddings. Both models use the Transformer\cite{vaswani2017attention} attention mechanism for representation learning. In ViLBERT, the image and text inputs are handled in two separate Transformers, which are later fused. In contrast, UNITER fuses the inputs in a single Transformer to model low and high-level features simultaneously. \cite{chen2020uniter} reports that UNITER outperforms ViLBERT with far fewer parameters in the downstream vision and language tasks.

\section{Problem Statement
\label{sec:problem}
}

\begin{figure}[t]
    \centering
    \includegraphics[width=\linewidth]{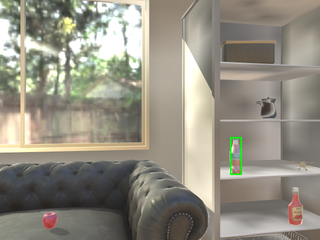}
    \caption{\small Typical scene of the MLU-FI task. The target object is enclosed with the green rectangle, given the instruction ``Pick up the empty bottle on the shelf.''}
    \label{fig:sample}
\end{figure}

\begin{figure*}[t]
    \centering
    \includegraphics[width=\linewidth]{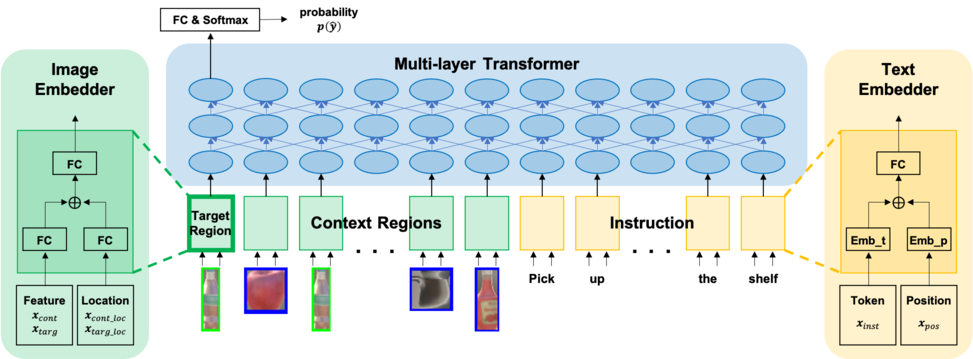}
    \caption{\small Proposed method framework: Target-dependent UNITER consists of Image Embedder, Text Embedder, and Multi-layer Transformer.}
    \label{fig:network}
\end{figure*}

In this paper, we define the task as MLU-FI to identify the target object given an instruction and context regions.
Fig.~\ref{fig:sample} shows a typical scene of the task. For this scene, we consider the instruction ``Pick up the empty bottle on the shelf.'' Here, the model is required to identify the empty bottle on the shelf as the target object.

The task is a binary classification that judges whether each object is the target object or not. Note that the task is not a multi-class classification, in which one single object is selected from among all of the objects in the image.
This allows us to take into account the case where there are multiple target objects in the image, and the case where the image does not contain any target objects.
Ideally, the model will output ``1'' if the candidate region contains the target object, and ``0'' otherwise.
MLU-FI is characterized by the following:
\begin{itemize}
    \item \textbf{Input}: An instruction, a target candidate region, and context regions.
    \item \textbf{Output}: A predicted probability that the target candidate is correct given the instruction.
\end{itemize}
The terminology used in this paper is defined as follows:
\begin{itemize}
    \item \textbf{Target object}: Object to be manipulated.
    \item \textbf{Target candidate}: Object to be judged to determine whether it is the target object or not.
    \item \textbf{Context regions}: All regions detected by an object detecter.
\end{itemize}
Context regions are extracted by inputting the whole image into Faster R-CNN\cite{ren2016faster}, which is an end-to-end object detection model. In Faster R-CNN, ResNet101\cite{he2016deep} is used to generate a feature map.
The evaluation metric is the classification accuracy.

\section{Proposed Method
\label{sec:proposed}
}

Fig.~\ref{fig:network} shows the framework of our method. In the figure, ``Instruction,'' ``Target Region,'' and ``Context Regions'' denote the instruction, the region of the target candidate, and the regions containing objects around the target object, respectively. Our method consists of three modules: Image Embedder, Text Embedder, and Multi-layer Transformer. The Text Embedder consists of two embedding layers and a normalization layer. The Image Embedder consists of two fully connected (FC) layers and a normalization layer.
Because the instruction and object regions are handled in the single Transformer, our model is expected to learn the relationships between each object in the image and each word in the instruction.

\subsection{Novelty}

The difference between our method and the baseline MTCM\cite{magassouba2019understanding} can be specified as follows:
\begin{itemize}
    \item While MTCM processes instructions and images separately through LSTM\cite{hochreiter1997long} and VGG16\cite{simonyan2014very}, our method processes them using a single Transformer\cite{vaswani2017attention}.
    \item Our method models the relationship between the target candidate and other objects, whereas MTCM does not.
    \item Our method is based on the UNITER\cite{chen2020uniter} framework. Unlike UNITER, our method considers the given information about the target candidate.
\end{itemize}

\subsection{Target-dependent UNITER}

UNITER is a generic pretraining model that can handle heterogeneous downstream vision and language tasks, which is advantageous for overcoming data shortages. In the pretraining stage, we perform the four tasks proposed in \cite{chen2020uniter}: masked language modeling, masked region modeling, image--text matching, and word--region alignment.

\subsubsection{Input}

The input $\bm{x}$ of Target-dependent UNITER is defined as follows:
\begin{align}
    \bm{x}&=\{\bm{X}_{inst},\bm{X}_{cont},\bm{X}_{targ}\}, \\
    \bm{X}_{inst}&=\{\bm{x}_{inst},\bm{x}_{pos}\}, \\
    \bm{X}_{targ}&=\{\bm{x}_{targ},\bm{x}_{targloc}\}, \\
    \bm{X}_{cont}&=\{(\bm{x}^{(i)}_{cont},\bm{x}^{(i)}_{contloc})|i=1,\dots, N\},
\end{align}
where $\bm{x}_{inst}$, $\bm{x}_{targ}$, $\bm{x}^{(i)}_{cont}$, $\bm{x}_{pos}$, $\bm{x}_{targloc}$, and $\bm{x}^{(i)}_{contloc}$ denote the instruction, target candidate region, context regions, positions of words in the instruction, location of target candidate region, and location of context regions, respectively. As stated above, the regions are detected by Faster R-CNN\cite{ren2016faster}. The number of regions is denoted as $N$.

\begin{figure}[t]
    \centering
    \includegraphics[width=\linewidth]{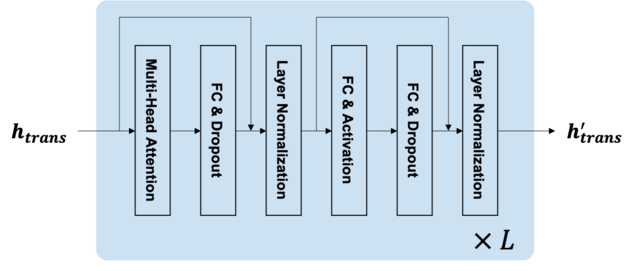}
    \caption{\small Multi-layer Transformer: a single Transformer layer consists of multi-head attention layer, fully connected layer, dropout layer, normalization layer, and activation function.}
    \label{fig:transformer}
\end{figure}

\subsubsection{Text Embedder}

The Text Embedder performs the embedding process for the instruction. Its input consists of $\bm{x}_{inst}$ and $\bm{x}_{pos}$. First, we tokenize the instruction using WordPiece and convert it into a token sequence.

Let $n_i$ denote the word index of the $i$-th token of the sequence, $\bm{x}_{inst}$ represents the one-hot vector set where the $n_i$-th element is 1, and $\bm{x}_{pos}$ represents the one-hot vector set where the $i$-th element is 1. We multiply these vectors by $W_{inst}$ and $W_{pos}$, respectively, to give the linear embedding. Note that the weights $W_{inst}$ and $W_{pos}$ are updated in the training stage. Finally, each output is concatenated and input to the FC layer $f_{\mathrm{FC}}(\cdot)$ to obtain the following output:
\begin{align}
    \bm{h}_{txtemb}^{\prime}=f_{\mathrm{FC}}(\{W_{inst}\bm{x}_{inst},W_{pos}\bm{x}_{pos}\}).
\end{align}

\subsubsection{Image Embedder}

The Image Embedder performs the embedding process for the target candidate region and context regions. Its input consists of $\bm{x}_{cont}^{(i)}$, $\bm{x}_{contloc}^{(i)}$, $\bm{x}_{targ}$, and $\bm{x}_{targloc}$. $\bm{x}_{cont}^{(i)}$ represents the feature vectors of each object region extracted from the image by Faster R-CNN. In Faster R-CNN, ResNet101\cite{he2016deep} is used to generate a feature map. $\bm{x}_{contloc}^{(i)}$ represents the vector of location features of each object region, and is a seven-dimensional vector $[x_1,y_1,x_2,y_2,w,h,{w}\times{h}]$ (normalized left/top/right/bottom coordinates, width, height, and area). Each of these vectors is input to the FC layer, and then concatenated and input to the FC layer again. The output $\bm{h}_{cont}^{\prime(i)}$ is obtained as
\begin{align}
    \bm{h}_{cont}^{\prime(i)}=f_{\mathrm{FC}}(\{f_{\mathrm{FC}}(\bm{x}_{cont}^{(i)}),f_{\mathrm{FC}}(\bm{x}_{contloc}^{(i)})\}).
\end{align}

$\bm{x}_{targ}$ and $\bm{x}_{targloc}$ are selected from the vectors $\{\bm{x}_{cont}^{(1)},\dots,\bm{x}_{cont}^{(N)}\}$ and $\{\bm{x}_{contloc}^{(1)},\dots,\bm{x}_{contloc}^{(N)}\}$. The output $\bm{h}_{targ}^{\prime}$ is obtained as
\begin{align}
    \bm{h}_{targ}^{\prime}=f_{\mathrm{FC}}(\{f_{\mathrm{FC}}(\bm{x}_{targ}),f_{\mathrm{FC}}(\bm{x}_{targloc})\}).
\end{align}
The final output $\bm{h}_{imgemb}^{\prime}$ is obtained as
\begin{align}
    \bm{h}_{imgemb}^{\prime}=\{\bm{h}_{targ}^{\prime},\bm{h}_{cont}^{\prime(1)},\dots,\bm{h}_{cont}^{\prime(N)}\},
\end{align}

\subsubsection{Multi-layer Transformer}

\begin{table}[t]
    \normalsize
    \caption{\small Parameters settings and structure of Target-dependent UNITER}
    \label{tab:param}
    \centering
    \begin{tabular}{|l|l|}
        \hline
        Transformer & \#L: 2, \#H: 768, \#A: 12 \\
        \hline
        Optimizer & AdamW \\ & $(\beta_{1}=0.9, \beta_{2}=0.98)$ \\
        \hline
        Learning rate & $8\times10^{-5}$ \\
        \hline
        Weight decay & 0.01 \\
        \hline
        Step & 20000 \\
        \hline
        Batch size & 8 \\
        \hline
        Dropout & 0.1 \\
        \hline
    \end{tabular}
\end{table}

\begin{table*}[t]
    \centering
    \begin{tabular}{c}
        \begin{minipage}{0.64\hsize}
            \normalsize
            \caption{\small Accuracy on the PFN-PIC and WRS-UniALT datasets}
            \label{tab:quantitive_result}
            \centering
            \begin{tabular}{l|c|c}
                \hline
                \multicolumn{1}{l|}{} & \multicolumn{2}{c}{Accuracy [\%]} \\
                \cline{2-3}
                Method & PFN-PIC & WRS-UniALT \\
                \hline
                \hline
                Baseline (MTCM\cite{magassouba2019understanding}) & $90.1\pm0.93$ & $91.8\pm0.36$ \\
                \hline
                (i) Ours (W/o FRCNN fine-tuning) & $91.5\pm0.69$ & $94.0\pm1.49$ \\
                \hline
                (ii) Ours (Late fusion) & $96.0\pm0.08$ & $96.0\pm0.24$ \\
                \hline
                (iii) Ours (Few context regions) & $96.6\pm0.36$ & $95.8\pm0.71$ \\
                \hline
                (iv) Ours (W/o pretraining) & $96.8\pm0.34$ & $95.4\pm0.19$ \\
                \hline
                Ours (Target-dependent UNITER) & $\mathbf{96.9\pm0.34}$ & $\mathbf{96.4\pm0.24}$ \\
                \hline
            \end{tabular}
        \end{minipage}
        \begin{minipage}{0.34\hsize}
            \normalsize
            \caption{\small Confusion matrix on the PFN-PIC and WRS-UniALT datasets}
            \label{tab:confusion_matrix}
            \centering
            \begin{tabular}{l|c|c}
                \hline
                & PFN-PIC & WRS-UniALT \\
                \hline
                \hline
                TP & 298 & 115 \\
                \hline
                TN & 296 & 108 \\
                \hline
                FP & 10 & 8 \\
                \hline
                FN & 8 & 1 \\
                \hline
            \end{tabular}
        \end{minipage}
    \end{tabular}
\end{table*}

Fig.~\ref{fig:transformer} shows the structure of the Multi-layer Transformer. This module consists of the following layers: (a) multi-head attention layer, (b) FC layer, (c) dropout layer, (d) normalization layer, (e) FC layer, (f) activation function, (g) FC layer, (h) dropout layer, and (i) normalization layer. We define (a)--(i) as a single Transformer layer.
The input $\bm{h}_{trans}$ is obtained as follows:
\begin{align}
    \bm{h}_{trans}=\{\bm{h}_{txtemb}^{\prime},\bm{h}_{imgemb}^{\prime}\}
\end{align}
First, we generate the queries $Q^{(i)}$, the keys $K^{(i)}$, and the values $V^{(i)}$ as follows:
\begin{align}
    Q^{(i)}&=W_q^{(i)}\bm{h}_{trans}^{(i)}, \\
    K^{(i)}&=W_k^{(i)}\bm{h}_{trans}^{(i)}, \\
    V^{(i)}&=W_v^{(i)}\bm{h}_{trans}^{(i)}.
\end{align}
Then, we calculate the attention score $S_{attn}$ as
\begin{align}
    S_{attn}&=\{f_{attn}^{(1)},\dots,f_{attn}^{(A)}\}, \\
    f_{attn}^{(i)}&=V^{(i)}\mathrm{softmax}(\frac{Q^{(i)}K^{(i)\top}}{\sqrt{d_k}}), \\
    d_k&=\frac{H}{A},
\end{align}
where $H$ and $A$ denote the hidden size and the number of attention heads, respectively.

The final output of the entire model $p(\hat{\bm{y}})$ is obtained as follows:
\begin{align}
    p(\hat{\bm{y}})=\mathrm{softmax}(f_{\mathrm{FC}}(\bm{h}_{trans}^{\prime})),
\end{align}
where $\bm{h}_{trans}^{\prime}$ and $\hat{\bm{y}}$ denote the output of the final Transformer layer and the prediction, respectively.

We used the following loss function:
\begin{align}
    \mathcal{L}(\hat{\bm{y}})=-\sum_{n}\sum_{j}y_{nj}\log p(\hat{y}_{nj})
\end{align}
where $y_{nj}$ denotes the label of the $j$-th dimension of the $n$-th sample and $\hat{y}_{nj}$ denotes the output of the $j$-th dimension of the $n$-th sample.

\section{Experiments
\label{sec:experiments}
}

\subsection{Dataset}

In the experiments, we evaluated our method on following two datasets.

\subsubsection{The PFN-PIC dataset}

PFN-PIC\cite{hatori2018interactively} is a standard dataset consisting of images and a set of instructions. Each image contains approximately 20 everyday items scattered in four boxes. Each instruction is an object-fetching instruction that requires one object to be moved to another box; all instructions are written in both Japanese and English. The instructions were collected using the Amazon Mechanical Turk platform, and at least three annotators were assigned to each object.
The PFN-PIC dataset contains 1180 images and 90759 sentences; the vocabulary size is 4682 words, the total number of words is 1293301, and the average sentence length is 14.2 words.
In our experiment, the training, validation, and test sets consisted of 63330, 710, and 612 samples, respectively, following the dataset balancing described below.

\subsubsection{The WRS-UniALT dataset}

We built an additional dataset WRS-UniALT, which is a simulation-based dataset consisting of images and a set of instructions. The images were collected in the standard simulator of World Robot Summit Partner Robot Challenge\cite{okada2019competitions} by DSRs. Each image contains approximately 5 everyday items located in a room. Each instruction is an object-fetching instruction written in English. We requested six annotators to create the instructions.
The WRS-UniALT dataset contains 570 images and 1246 sentences, with a vocabulary size of 167 words, a total of 8816 words, and an average sentence length of 7.1 words.
In this experiment, the training, validation, and test sets consisted of 2048, 210, and 232 samples, respectively, following the dataset balancing described below.

\subsubsection{Dataset preprocessing}

In the preprocessing stage for our method, we extracted object regions from each image in the datasets using Faster R-CNN\cite{ren2016faster}. However, each detected region did not always match the ground truth. Therefore, we divided those regions into positive and negative samples based on the intersection over union (IoU), $\beta$. Samples with $\beta>0.7$ and $\beta<0.3$ were labeled as positive and negative, respectively. In addition, we reduced the number of negative samples so as to balance the numbers of positive and negative samples.

\subsection{Experimental Setup}

The experimental setup is summarized in Table~\ref{tab:param}, where \#L denotes the number of layers, \#H is the hidden size, \#A is the number of attention heads in the Transformer\cite{vaswani2017attention}, and each step represents the processing of one batch.

Target-dependent UNITER had 42 M parameters for pretraining and 39 M parameters for fine-tuning.
The parameters were trained on an RTX 2080 with 11GB of GPU memory and an Intel Core i9 processor.
Target-dependent UNITER required 3 h for pretraining and 1 h for fine-tuning on MLU-FI.
We evaluated our model on the validation and test sets every 2000 training steps. The final performance was given by the test set accuracy when the validation set accuracy was maximized.

\subsection{Quantitative Results}

\begin{figure*}[t]
    \centering
    \begin{tabular}{c}
        \begin{minipage}{0.24\hsize}
            \centering
            \includegraphics[width=\linewidth]{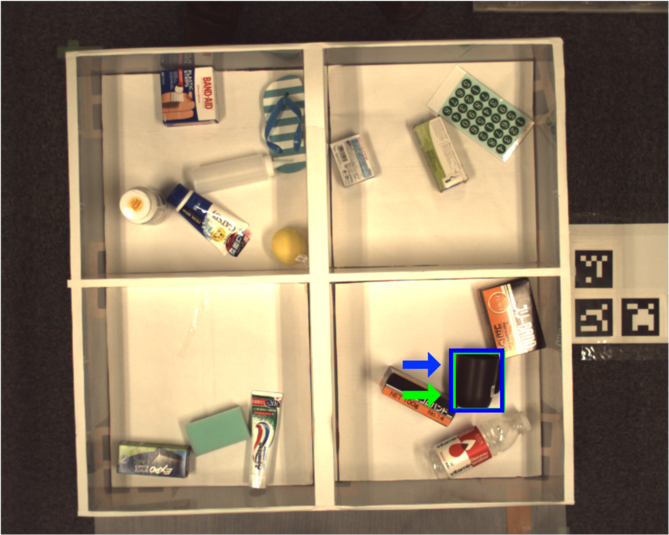}
            \hspace{1.6cm}(\small a)
        \end{minipage}
        \begin{minipage}{0.24\hsize}
            \centering
            \includegraphics[width=\linewidth]{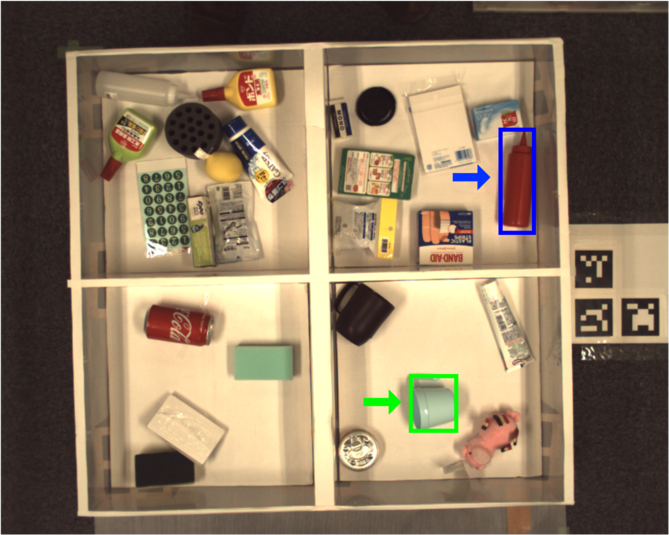}
            \hspace{1.6cm}(\small b)
        \end{minipage}
        \begin{minipage}{0.24\hsize}
            \centering
            \includegraphics[width=\linewidth]{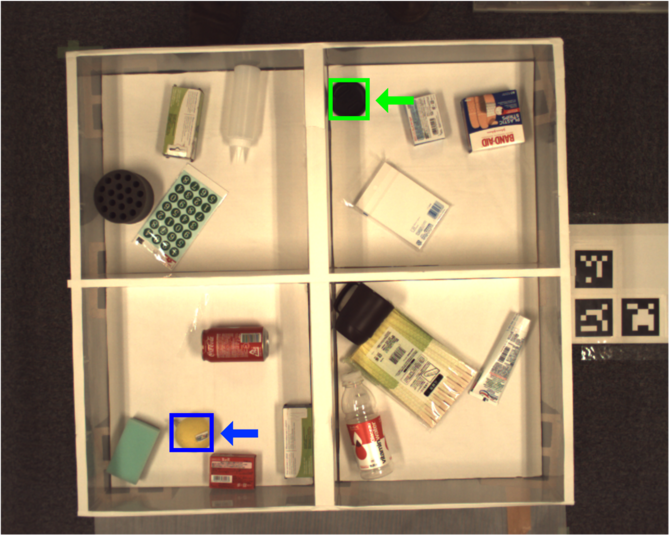}
            \hspace{1.6cm}(\small c)
        \end{minipage}
        \begin{minipage}{0.24\hsize}
            \centering
            \includegraphics[width=\linewidth]{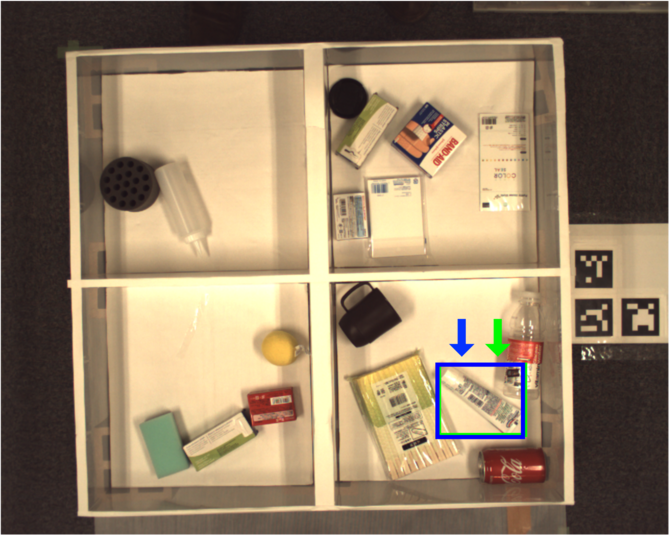}
            \hspace{1.6cm}(\small d)
        \end{minipage}
    \end{tabular}
    \caption{Qualitative results on the PFN-PIC dataset. The prediction is given in blue while the ground truth is in green. Panels (a), (b), (c), and (d) show TP, TN, FP, and FN samples, respectively.
    (a) ``Pick up the black cup in the bottom right section of the box and move it to the bottom left section of the box.'' (b) ``Grab the sky blue cup and put it in the upper right box.'' (c) ``Move the black circular object on the top right to the bottom right.'' (d) ``Grab the small white plastic container on the lower right box to the upper left box.''}
    \label{fig:qualitative_result_pfn}
\end{figure*}

\begin{figure*}[t]
    \centering
    \begin{tabular}{c}
        \begin{minipage}{0.24\hsize}
            \centering
            \includegraphics[width=\linewidth]{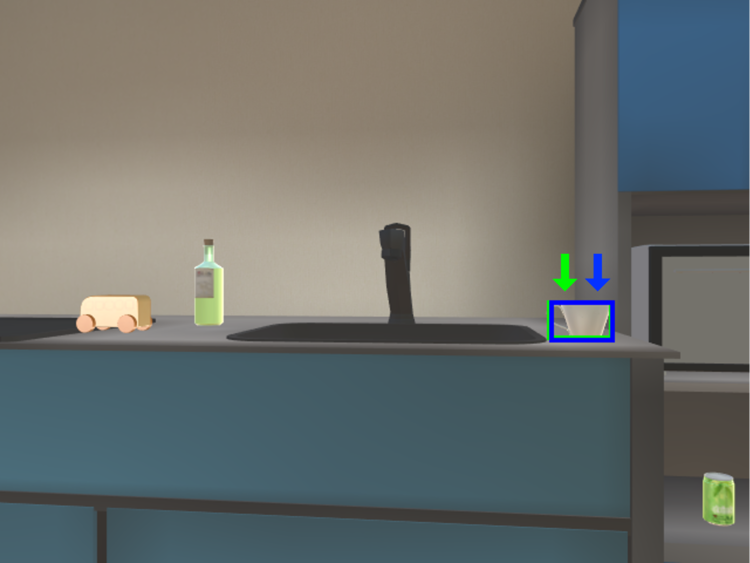}
            \hspace{1.6cm}(\small a)
        \end{minipage}
        \begin{minipage}{0.24\hsize}
            \centering
            \includegraphics[width=\linewidth]{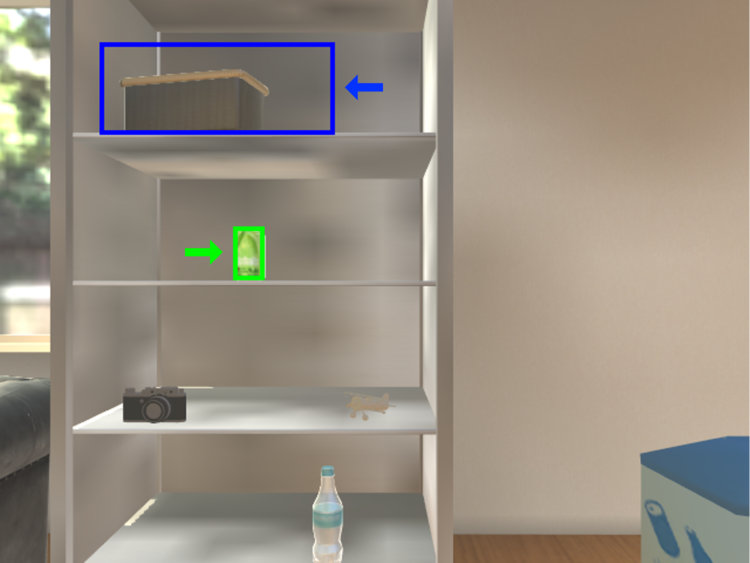}
            \hspace{1.6cm}(\small b)
        \end{minipage}
        \begin{minipage}{0.24\hsize}
            \centering
            \includegraphics[width=\linewidth]{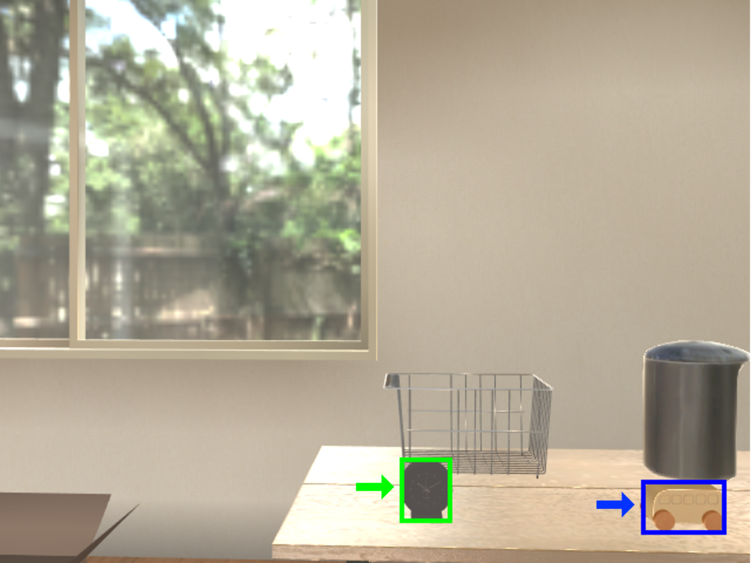}
            \hspace{1.6cm}(\small c)
        \end{minipage}
        \begin{minipage}{0.24\hsize}
            \centering
            \includegraphics[width=\linewidth]{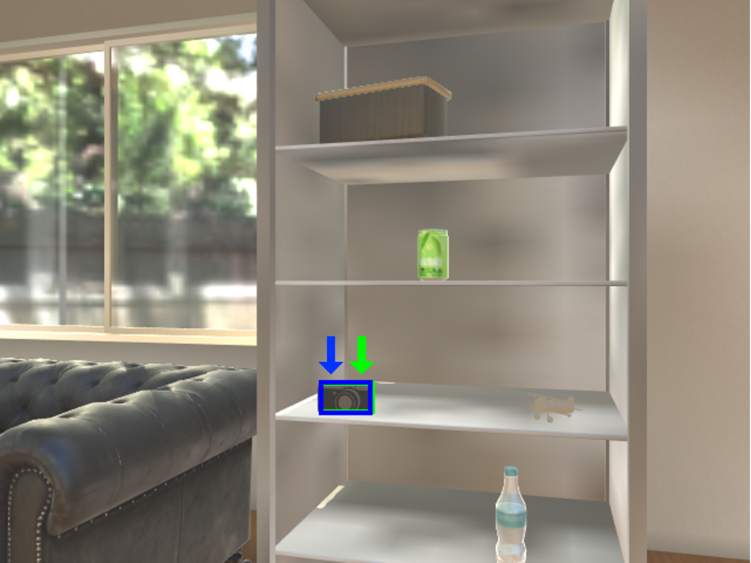}
            \hspace{1.6cm}(\small d)
        \end{minipage}
    \end{tabular}
    \caption{Qualitative results on the WRS-UniALT dataset. The prediction is given in blue while the ground truth is in green. Panels (a), (b), (c), and (d) show TP, TN, FP, and FN samples, respectively.
    (a) ``Give me the white cup.'' (b) ``Take the can juice on the white shelf.'' (c) ``Go get me the black clock on the table.'' (d) ``Pick up the camera near the can juice.''}
    \label{fig:qualitative_result_wrs}
\end{figure*}

Table~\ref{tab:quantitive_result} presents the quantitative results.
The evaluation metric is the classification accuracy $Acc$, given by
\begin{align}
    Acc=\frac{TP+TN}{TP+FP+FN+TN},
\end{align}
where $TP$, $FP$, $FN$, and $TN$ denote true positive, false positive, false negative, and true negative, respectively.
Because there are equal numbers of positive and negative samples in the datasets, the chance performance is 50\%.

Table~\ref{tab:quantitive_result} shows that Target-dependent UNITER achieved accuracy of 96.9\% on the PFN-PIC dataset, while the baseline method achieved 90.1\%. Target-dependent UNITER achieved accuracy of 96.4\% on the WRS-UniALT dataset, while the baseline method achieved 91.8\%. The results by MTCM were reproduced by ourselves. They were slightly lower than those reported in \cite{magassouba2019understanding} because the source information was not used as an auxiliary output labels for a fair comparison. Target-dependent UNITER outperformed the MTCM by 6.5\% and 4.6\% on the PFN-PIC and WRS-UniALT datasets, respectively.

Table~\ref{tab:confusion_matrix} presents the confusion matrix for our method.
On the PFN-PIC dataset, there were 298, 296, 10, and 8 samples classified as TP, TN, FP, and FN, respectively. Thus, there were 18 samples that constitute failure cases.
On the WRS-UniALT dataset, there were 115, 108, 8, and 1 samples classified as TP, TN, FP, and FN, respectively, giving 9 failure cases. These failure cases are further analysed in Section.~\ref{sec:error_analysis}.

\subsection{Ablation Studies}

We set the following four conditions as the ablation study:
\begin{enumerate}
    \renewcommand{\labelenumi}{(\roman{enumi})}
    \item W/o FRCNN fine-tuning: In this model, Faster R-CNN was not fine-tuned on each dataset to investigate the effect of the accuracy of the object detection.
    \item Late fusion: In this model, the structure of our model was changed to investigate the difference in performance between ``early fusion'' and ``late fusion.'' Our method has an early fusion structure. In the early fusion structure, the target region, the context regions, and the instruction are integratedly handled by the single Transformer network. In the late fusion structure, the target region and the other inputs are handled by separate networks. Their outputs are fused at the end of the network.
    \item Few context regions: In this model, the number of context regions was reduced by half to investigate the effect of the number of context regions.
    \item W/o pretraining: In this model, the pretraining of UNITER was removed to investigate the difference in performance with and without the pretraining.
\end{enumerate}
Table~\ref{tab:quantitive_result} shows that the accuracy on the PFN-PIC dataset decreased 5.4\%, 0.9\%, 0.3\%, and 0.1\% in the condition (i), (ii), (iii), and (iv), respectively. It also shows that the accuracy on the WRS-UniALT dataset decreased 2.4\%, 0.4\%, 0.6\%, and 1.0\% in the condition (i), (ii), (iii), and (iv), respectively.
On the PFN-PIC dataset, the largest contribution was the fine-tuning of Faster R-CNN and the second largest contribution was the early fusion. On the WRS-UniALT dataset, the largest contribution was the fine-tuning of Faster R-CNN and the second largest contribution was the pretraining. These results indicate that the pretraining and the early fusion of Target-dependent UNITER were beneficial to the performance. Note that the improvement of the accuracy of Faster R-CNN is out of scope.

\subsection{Qualitative Results}

\subsubsection{PFN-PIC qualitative results}

Fig.~\ref{fig:qualitative_result_pfn} shows the qualitative results on the PFN-PIC dataset.

Fig.~\ref{fig:qualitative_result_pfn} (a) shows a TP sample. The target object was the black cup in the lower right box, and the detected region represented almost the same region as the ground truth. The model output $p(\hat{\bm{y}})=0.999$ and correctly identified the target candidate as the target object.
Fig.~\ref{fig:qualitative_result_pfn} (b) shows a TN sample. The target object was the blue cup in the lower right box, and the detected region represented a different region from the ground truth. The model output $p(\hat{\bm{y}})=2.37\times10^{-9}$ and correctly identified the target candidate as a different object from the target object.

Fig.~\ref{fig:qualitative_result_pfn} (c) shows an FP sample. The target object was the black object in the upper right box, and the detected region represented a different region from the ground truth. The model output $p(\hat{\bm{y}})=0.999$ and mistakenly identified the target candidate as the target object.
Fig.~\ref{fig:qualitative_result_pfn} (d) shows an FN sample. The target object was the white tube in the lower right box, and the detected region represented almost the same region as the ground truth. The model output $p(\hat{\bm{y}})=2.93\times10^{-3}$ and mistakenly identified the target candidate as a different object from the target object.
The causes of above errors are analysed in Section.~\ref{sec:error_analysis}.

\subsubsection{WRS-UniALT qualitative results}

Fig.~\ref{fig:qualitative_result_wrs} shows the qualitative results on the WRS-UniALT dataset.

Fig.~\ref{fig:qualitative_result_wrs} (a) shows a TP sample. The target object was the white cup on the sink, and the detected region represented almost the same region as the ground truth. The model output $p(\hat{\bm{y}})=0.999$ and correctly identified the target candidate as the target object.
Fig.~\ref{fig:qualitative_result_wrs} (b) shows a TN sample. The target object was the green canned juice on the shelf, and the detected region represented a different region from the ground truth. The model output $p(\hat{\bm{y}})=8.19\times10^{-18}$ and correctly identified the target candidate as a different object from the target object.

Fig.~\ref{fig:qualitative_result_wrs} (c) shows an FP sample. The target object was the black clock on the table, and the detected region represented a different region from the ground truth. The model output $p(\hat{\bm{y}})=0.999$ and mistakenly identified the target candidate as the target object.
Fig.~\ref{fig:qualitative_result_wrs} (d) shows an FN sample. The target object was the black camera on the shelf, and the detected region represented almost the same region as the ground truth. The model output $p(\hat{\bm{y}})=2.13\times10^{-9}$ and mistakenly identified the target candidate as a different object from the target object.
The causes of above errors are analysed in Section.~\ref{sec:error_analysis}.

\subsection{Error Analysis
\label{sec:error_analysis}
}

\begin{table}[t]
    \normalsize
    \caption{\small Categorization of the erroneous predictions on the PFN-PIC and WRS-UniALT datasets}
    \label{tab:categorization}
    \centering
    \begin{tabular}{l|l|c}
        \hline
        Error ID & Description & \#Error \\
        \hline
        \hline
        ER & Excessive region & 8 \\
        \hline
        AE & Ambiguous expression & 4 \\
        \hline
        OVW & Out-of-vocabulary word & 4 \\
        \hline
        IVI & Insufficient visual information & 3 \\
        \hline
        OCE & Other comprehension error & 8 \\
        \hline
        \hline
        Total & - & 27 \\
        \hline
    \end{tabular}
\end{table}

\begin{figure}[t]
    \centering
    \begin{tabular}{c}
        \begin{minipage}{0.48\hsize}
            \centering
            \includegraphics[width=\linewidth]{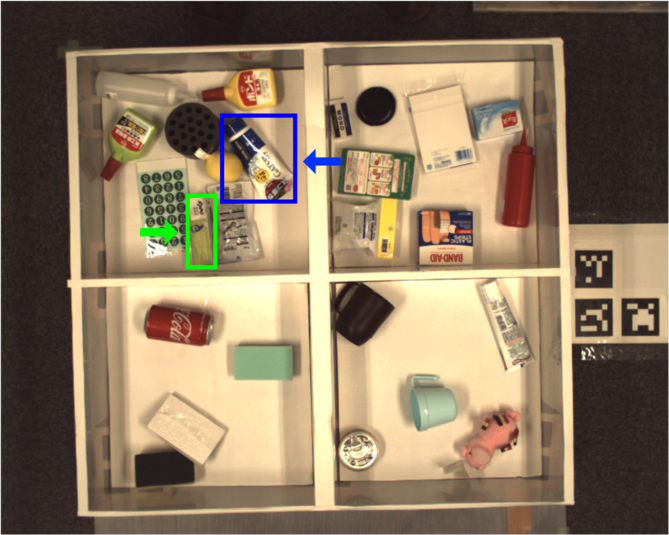}
            \vspace{2mm}(\small a)
            \includegraphics[width=\linewidth]{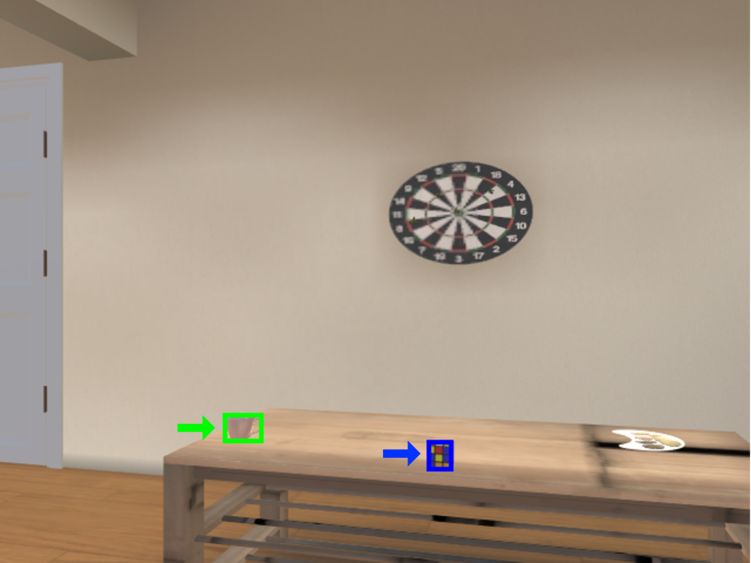}
            (\small c)
        \end{minipage}
        \begin{minipage}{0.48\hsize}
            \centering
            \includegraphics[width=\linewidth]{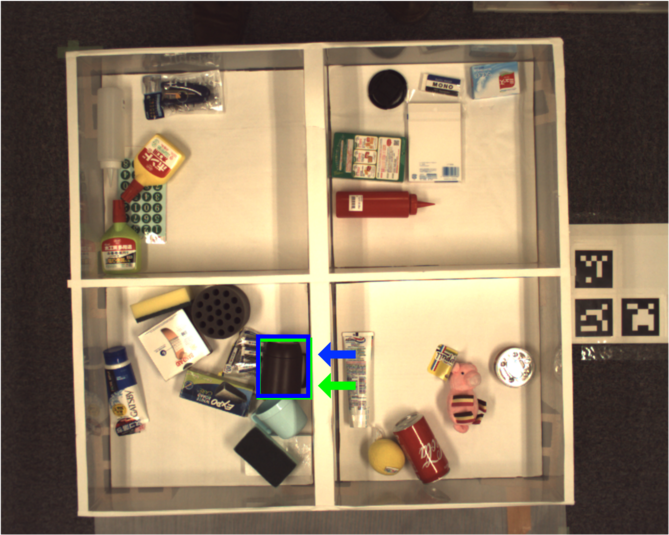}
            \vspace{2mm}(\small b)
            \includegraphics[width=\linewidth]{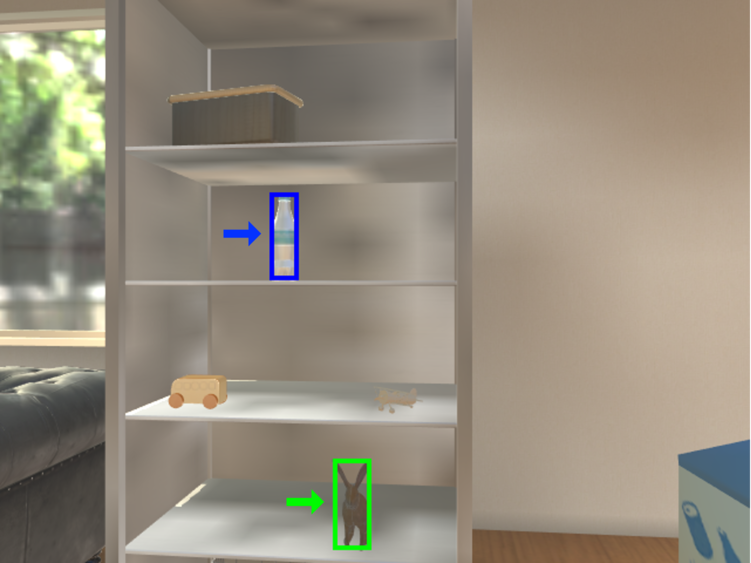}
            (\small d)
        \end{minipage}
    \end{tabular}
    \caption{Samples of failure cases. The prediction is given in blue while the ground truth is in green. Panels (a) and (b) show ER and AE samples, respectively. Panels (c) and (d) show IVI samples.
    (a) ``Move the green rectangle with white on the side from the upper left box, to the lower left box.'' (b) ``Grab black object in the middle right of the bottom left box and put in top right box.'' (c) ``Take the white cup on the corner of the table.'' (d) ``Bring me the rabbit doll on the lower part of the shelf.''}
    \label{fig:failure_cases}
\end{figure}

Table~\ref{tab:categorization} categorizes the failure cases and Fig.~\ref{fig:failure_cases} shows typical samples. The causes of failure can be roughly divided into five groups:
\begin{itemize}
    \item Excessive region (ER): The region contains the features of multiple objects because the object of interest is placed diagonally in the image. Fig.~\ref{fig:failure_cases} (a) shows an ER sample, in which the detected region contains the objects around the ``blue tube'' target object.
    \item Ambiguous expression (AE): The instruction includes ambiguous expressions about the location of the target object, which makes it difficult to select a single target object. Fig.~\ref{fig:failure_cases} (b) shows an AE sample, in which the location of the target object is given as ``in the middle right,'' but this is too ambiguous for the model to specify the target.
    \item Out-of-vocabulary word (OOV): The instruction contains out-of-vocabulary words or a tokenization error occurs as the result of the limited vocabulary size. There was an OOV sample with the instruction ``Grab the white rectangle small box and put it in the lower right box,'' in which WordPiece tokenized ``rectangle'' as (``re'', ``\#\#ct'', ``\#\#ang'', ''\#\#le'') and the model failed to comprehend the shape of the target object.
    \item Insufficient visual information (IVI): The region does not contain sufficient visual information on the target object because of its size or orientation. Fig.~\ref{fig:failure_cases} (c) shows a sample of the former, in which the ``white cup'' is too small for its features to be extracted. Fig.~\ref{fig:failure_cases} (d) shows a sample of the latter, in which the ``rabbit doll'' is facing straight at the camera and the model cannot recognize the inherent rabbit shape.
    \item Other comprehension error (OCE): This contains failure cases that cannot be categorized into the above groups.
\end{itemize}
Table~\ref{tab:categorization} presents that the major errors by our method are related to ERs that were often caused by the shape of regions. It is considered that we can reduce ERs by using a semantic segmentation method instead of an object detection with rectangle regions.

\section{Conclusion
\label{sec:conclusion}
}

In this paper, we have described the Target-dependent UNITER method for understanding object-fetching instructions that contain referring expressions about the target object.
We would like to emphasize the following contributions of this study:
\begin{itemize}
    \item We introduced the Transformer\cite{vaswani2017attention} attention mechanism based on the UNITER\cite{chen2020uniter} framework into the field of object-manipulation instruction comprehension. The proposed model can be pretrained to learn the relationship between objects and sentence.
    \item We extended the UNITER framework by introducing a new structure for handling target candidates.
    \item Target-dependent UNITER outperformed the baseline MTCM\cite{magassouba2019understanding} in terms of classification accuracy on two datasets.
\end{itemize}
In future work, we plan to apply this model to physical robots.

\bibliographystyle{IEEEtran}
\bibliography{reference}

\begin{thebibliography}{10}
\providecommand{\url}[1]{#1}
\csname url@rmstyle\endcsname
\providecommand{\newblock}{\relax}
\providecommand{\bibinfo}[2]{#2}
\providecommand\BIBentrySTDinterwordspacing{\spaceskip=0pt\relax}
\providecommand\BIBentryALTinterwordstretchfactor{4}
\providecommand\BIBentryALTinterwordspacing{\spaceskip=\fontdimen2\font plus
\BIBentryALTinterwordstretchfactor\fontdimen3\font minus
  \fontdimen4\font\relax}
\providecommand\BIBforeignlanguage[2]{{%
\expandafter\ifx\csname l@#1\endcsname\relax
\typeout{** WARNING: IEEEtran.bst: No hyphenation pattern has been}%
\typeout{** loaded for the language `#1'. Using the pattern for}%
\typeout{** the default language instead.}%
\else
\language=\csname l@#1\endcsname
\fi
#2}}

\bibitem{chen2020uniter}
Y.-C. Chen, L.~Li, L.~Yu, A.~El~Kholy, F.~Ahmed, Z.~Gan, Y.~Cheng, and J.~Liu,
  ``Uniter: Universal image-text representation learning,'' in \emph{European
  Conference on Computer Vision}.\hskip 1em plus 0.5em minus 0.4em\relax
  Springer, 2020, pp. 104--120.

\bibitem{magassouba2019understanding}
A.~Magassouba, K.~Sugiura, A.~T. Quoc, and H.~Kawai, ``Understanding natural
  language instructions for fetching daily objects using gan-based multimodal
  target--source classification,'' \emph{IEEE Robotics and Automation Letters},
  vol.~4, no.~4, pp. 3884--3891, 2019.

\bibitem{vaswani2017attention}
A.~Vaswani, N.~Shazeer, N.~Parmar, J.~Uszkoreit, L.~Jones, A.~N. Gomez,
  {\L}.~Kaiser, and I.~Polosukhin, ``Attention is all you need,'' in
  \emph{Advances in neural information processing systems}, 2017, pp.
  5998--6008.

\bibitem{mogadala2020trends}
A.~Mogadala, M.~Kalimuthu, and D.~Klakow, ``Trends in integration of vision and
  language research: A survey of tasks, datasets, and methods,'' 2020.

\bibitem{antol2015vqa}
S.~Antol, A.~Agrawal, J.~Lu, M.~Mitchell, D.~Batra, C.~L. Zitnick, and
  D.~Parikh, ``Vqa: Visual question answering,'' in \emph{Proceedings of the
  IEEE international conference on computer vision}, 2015, pp. 2425--2433.

\bibitem{harabagiu2000experiments}
S.~Harabagiu, M.~Pasca, and S.~J. Maiorano, ``Experiments with open-domain
  textual question answering,'' in \emph{COLING 2000 Volume 1: The 18th
  International Conference on Computational Linguistics}, 2000.

\bibitem{krahmer2012computational}
E.~Krahmer and K.~Van~Deemter, ``Computational generation of referring
  expressions: A survey,'' \emph{Computational Linguistics}, vol.~38, no.~1,
  pp. 173--218, 2012.

\bibitem{thomas2014meaning}
J.~A. Thomas, \emph{Meaning in interaction: An introduction to
  pragmatics}.\hskip 1em plus 0.5em minus 0.4em\relax Routledge, 2014.

\bibitem{condoravdi2003entailment}
C.~Condoravdi, D.~Crouch, V.~De~Paiva, R.~Stolle, and D.~Bobrow, ``Entailment,
  intensionality and text understanding,'' in \emph{Proceedings of the
  HLT-NAACL 2003 workshop on Text meaning}, 2003, pp. 38--45.

\bibitem{bowman2015large}
S.~R. Bowman, G.~Angeli, C.~Potts, and C.~D. Manning, ``A large annotated
  corpus for learning natural language inference,'' \emph{arXiv preprint
  arXiv:1508.05326}, 2015.

\bibitem{malinowski2014multi}
M.~Malinowski and M.~Fritz, ``A multi-world approach to question answering
  about real-world scenes based on uncertain input,'' \emph{Advances in neural
  information processing systems}, vol.~27, pp. 1682--1690, 2014.

\bibitem{agrawal2018don}
A.~Agrawal, D.~Batra, D.~Parikh, and A.~Kembhavi, ``Don't just assume; look and
  answer: Overcoming priors for visual question answering,'' in
  \emph{Proceedings of the IEEE Conference on Computer Vision and Pattern
  Recognition}, 2018, pp. 4971--4980.

\bibitem{yang2019cross}
S.~Yang, G.~Li, and Y.~Yu, ``Cross-modal relationship inference for grounding
  referring expressions,'' in \emph{Proceedings of the IEEE Conference on
  Computer Vision and Pattern Recognition}, 2019, pp. 4145--4154.

\bibitem{luo2017comprehension}
R.~Luo and G.~Shakhnarovich, ``Comprehension-guided referring expressions,'' in
  \emph{Proceedings of the IEEE Conference on Computer Vision and Pattern
  Recognition}, 2017, pp. 7102--7111.

\bibitem{ogura2020alleviating}
T.~Ogura, A.~Magassouba, K.~Sugiura, T.~Hirakawa, T.~Yamashita, H.~Fujiyoshi,
  and H.~Kawai, ``Alleviating the burden of labeling: Sentence generation by
  attention branch encoder--decoder network,'' \emph{IEEE Robotics and
  Automation Letters}, vol.~5, no.~4, pp. 5945--5952, 2020.

\bibitem{xie2019visual}
N.~Xie, F.~Lai, D.~Doran, and A.~Kadav, ``Visual entailment: A novel task for
  fine-grained image understanding,'' \emph{arXiv preprint arXiv:1901.06706},
  2019.

\bibitem{hochreiter1997long}
S.~Hochreiter and J.~Schmidhuber, ``Long short-term memory,'' \emph{Neural
  computation}, vol.~9, no.~8, pp. 1735--1780, 1997.

\bibitem{simonyan2014very}
K.~Simonyan and A.~Zisserman, ``Very deep convolutional networks for
  large-scale image recognition,'' \emph{arXiv preprint arXiv:1409.1556}, 2014.

\bibitem{magassouba2020multimodal_01}
A.~Magassouba, K.~Sugiura, and H.~Kawai, ``A multimodal target-source
  classifier with attention branches to understand ambiguous instructions for
  fetching daily objects,'' \emph{IEEE Robotics and Automation Letters},
  vol.~5, no.~2, pp. 532--539, 2020.

\bibitem{fukui2019attention}
H.~Fukui, T.~Hirakawa, T.~Yamashita, and H.~Fujiyoshi, ``Attention branch
  network: Learning of attention mechanism for visual explanation,'' in
  \emph{Proceedings of the IEEE Conference on Computer Vision and Pattern
  Recognition}, 2019, pp. 10\,705--10\,714.

\bibitem{hatori2018interactively}
J.~Hatori, Y.~Kikuchi, S.~Kobayashi, K.~Takahashi, Y.~Tsuboi, Y.~Unno, W.~Ko,
  and J.~Tan, ``Interactively picking real-world objects with unconstrained
  spoken language instructions,'' in \emph{2018 IEEE International Conference
  on Robotics and Automation (ICRA)}.\hskip 1em plus 0.5em minus 0.4em\relax
  IEEE, 2018, pp. 3774--3781.

\bibitem{kazemzadeh2014referitgame}
S.~Kazemzadeh, V.~Ordonez, M.~Matten, and T.~Berg, ``Referitgame: Referring to
  objects in photographs of natural scenes,'' in \emph{Proceedings of the 2014
  conference on empirical methods in natural language processing (EMNLP)},
  2014, pp. 787--798.

\bibitem{de2017guesswhat}
H.~De~Vries, F.~Strub, S.~Chandar, O.~Pietquin, H.~Larochelle, and
  A.~Courville, ``Guesswhat?! visual object discovery through multi-modal
  dialogue,'' in \emph{Proceedings of the IEEE Conference on Computer Vision
  and Pattern Recognition}, 2017, pp. 5503--5512.

\bibitem{liu2019clevr}
R.~Liu, C.~Liu, Y.~Bai, and A.~L. Yuille, ``Clevr-ref+: Diagnosing visual
  reasoning with referring expressions,'' in \emph{Proceedings of the IEEE
  Conference on Computer Vision and Pattern Recognition}, 2019, pp. 4185--4194.

\bibitem{okada2019competitions}
H.~Okada, T.~Inamura, and K.~Wada, ``What competitions were conducted in the
  service categories of the world robot summit?'' \emph{Advanced Robotics},
  vol.~33, no.~17, pp. 900--910, 2019.

\bibitem{lu2019vilbert}
J.~Lu, D.~Batra, D.~Parikh, and S.~Lee, ``Vilbert: Pretraining task-agnostic
  visiolinguistic representations for vision-and-language tasks,'' in
  \emph{Advances in Neural Information Processing Systems}, 2019, pp. 13--23.

\bibitem{ren2016faster}
S.~Ren, K.~He, R.~Girshick, and J.~Sun, ``Faster r-cnn: Towards real-time
  object detection with region proposal networks,'' \emph{IEEE transactions on
  pattern analysis and machine intelligence}, vol.~39, no.~6, pp. 1137--1149,
  2016.

\bibitem{he2016deep}
K.~He, X.~Zhang, S.~Ren, and J.~Sun, ``Deep residual learning for image
  recognition,'' in \emph{Proceedings of the IEEE conference on computer vision
  and pattern recognition}, 2016, pp. 770--778.

\end{thebibliography}

\end{document}